\documentclass[runningheads]{llncs}

\usepackage{eccv}

\usepackage{eccvabbrv}

\usepackage{graphicx}
\usepackage{booktabs}

\usepackage{algorithm}
\usepackage{algorithmic}
\usepackage{bm}
\usepackage{multirow}

\usepackage{float} 

\usepackage[accsupp]{axessibility}  

\usepackage{hyperref}

\usepackage{orcidlink}

\begin{document}

\title{PoseCrafter: One-Shot Personalized Video Synthesis Following Flexible Pose Control} 

\titlerunning{PoseCrafter: One-Shot Personalized Video Synthesis Following Flexible Poses}

\author{Yong Zhong\thanks{Equal contribution.}\inst{1} \and
Min Zhao$^\star$\inst{1} \and
Zebin You\inst{1} \and
Xiaofeng Yu\inst{2} \and
Changwang Zhang\inst{2} \and
Chongxuan Li\thanks{Correspondence to Chongxuan Li.}\inst{1}}

\authorrunning{Y.~Zhong et al.}

\institute{Gaoling School of AI, Renmin University of China, Beijing, China 
\and
Huawei Technologies Co., Ltd
\\
\email{yongzhong@ruc.edu.cn,\ gracezhao1997@gmail.com,\ zebin@ruc.edu.cn,\\ yuxiaofeng16@huawei.com,\ changwangzhang@foxmail.com,\\chongxuanli@ruc.edu.cn}
}

\maketitle

\begin{abstract}

  In this paper, we introduce PoseCrafter, a one-shot method for personalized video generation following the control of flexible poses. Built upon Stable Diffusion and ControlNet, we carefully design an inference process to produce high-quality videos without the corresponding ground-truth frames. First, we select an appropriate reference frame from the training video and invert it to initialize all latent variables for generation. Then, we insert the corresponding training pose into the target pose sequences to enhance faithfulness through a trained temporal attention module. Furthermore, to alleviate the face and hand degradation resulting from discrepancies between poses of training videos and inference poses, we implement simple latent editing through an affine transformation matrix involving facial and hand landmarks. Extensive experiments on several datasets demonstrate that PoseCrafter achieves superior results to baselines pre-trained on a vast collection of videos under 8 commonly used metrics.
 Besides, PoseCrafter can follow poses from different individuals or artificial edits and simultaneously retain the human identity in an open-domain training video. Our project page is available at \url{https://ml-gsai.github.io/PoseCrafter-demo/}.
\end{abstract}

\section{Introduction}
\label{sec:introduction}

The generation of digital humans, particularly through pose-guided techniques that allow for the creation of personalized human videos following specific pose instructions, holds significant promise in both practical and entertainment arenas. 
At the front of this area, pose-guided image-to-video approaches~\cite{xu2023magicanimate,chang2023magicdance,wang2023disco,hu2023animate} involve the conversion of a static human photograph into a dynamic video sequence adhering to given pose sequence. Despite its promising prospects, this technique confronts challenges due to the high-dimensional nature of video data, necessitating extensive datasets of high-quality human videos, often beyond public accessibility~\cite{hu2023animate}. Moreover, the computational costs involved in processing such data can be exceptionally daunting.

To circumvent the challenges associated with large-scale data requisites and substantial training expenditures, this paper focuses on a one-shot setting where we aim to generate videos, retaining the identity in a single training video and following the control of flexible poses. 
Notably, while existing video editing methods~\cite{wu2023tune,zhao2023controlvideo,liu2023video,shin2023edit,chai2023stablevideo} have provided valuable insights, their applicability is limited because they significantly rely on real video frames corresponding to the target poses. 
This prerequisite poses a significant challenge within the realm of video generation, where such frames are unavailable, compounding the difficulty of generating high-quality content. Particularly, our empirical findings, as detailed in ~\cref{tab:quantitative_results_TikTok} and \cref{tab:quantitative_results} vividly demonstrate that a representative approach~\cite{zhao2023controlvideo} produces videos with notably inferior quality.

To relieve this dependency, in this paper, 
we present PoseCrafter, a one-shot method for personalized video generation following the control of flexible poses. 
Built upon Stable Diffusion~\cite{rombach2022high} and ControlNet~\cite{zhang2023adding}, PoseCrafter requires merely fine-tuning the pre-trained model on a single open-domain video.
Technically, we carefully design an inference process to produce high-quality videos following flexible pose control without the corresponding frames. First, we select an appropriate reference frame from the training video and invert it to initialize all latent variables for generation. Then, we insert the corresponding training pose into the target pose sequences to enhance faithfulness through a trained temporal attention module.  Furthermore, to alleviate the face and hand degradation resulting from discrepancies between poses of training videos and inference poses, we implement simple latent editing through an affine transformation matrix involving facial and hand landmarks.

Compared with the aforementioned image-to-video methods~\cite{xu2023magicanimate,chang2023magicdance,wang2023disco,hu2023animate}, PoseCrafter is both computational and data efficient. Besides, leveraging a video as a source offers a richer tapestry of information than a single image, e.g., uncovering occluded details. Moreover, fine-tuning with a video source facilitates a deeper comprehension of individual identities and environmental nuances, significantly enhancing the authenticity and fidelity of the generated video content.

Experimentally, we conduct extensive evaluations on two publicly accessible datasets, TikTok~\cite{jafarian2021learning} and TED~\cite{siarohin2021motion}, covering real-life dancing and speaking scenarios. 
Quantitative (under 8 common metrics) and qualitative results (see \cref{sec:main_results}) demonstrate that PoseCrafter produces videos of higher quality than available image-to-video baselines~\cite{wang2023disco,xu2023magicanimate}, including commercial-grade applications such as GEN-2~\cite{esser2023structure}, and fine-tuning based approaches for specific subjects~\cite{wang2023disco,zhao2023controlvideo}. Moreover, we illustrate that PoseCrafter can follow flexible pose control including different poses from the same individual, artificially designed poses, and poses of different individuals (see \cref{sec:applications}). These findings strongly support the potential of one-shot methods for pose-guided video generation.

The key contributions of this paper are summarized as follows:

\begin{itemize}
    \item[\textbullet] We present PoseCrafter, a one-shot pose-guided personalized video generation framework, designed to circumvent the need for large-scale high-quality human video datasets and extensive training resources.
    \item[\textbullet] PoseCrafter introduces reference-frame selection and insertion to improve sample quality, and further alleviates face and hand degradation through simple latent editing.
    \item[\textbullet] PoseCrafter fine-tuned on a single video achieves superior results to baselines pre-trained on a vast collection of videos under 8 commonly used metrics.
    \item[\textbullet] PoseCrafter can follow poses from different individuals or artificial edits and simultaneously retain the human identity in an open-domain training video. 
\end{itemize}

\section{Related Work}
\label{sec:related_work}

\subsection{Controllable Image Generation}
Recently, because of the capacity to produce high-fidelity samples, diffusion models are inherently utilized in controllable image generation~\cite{zhao2022egsde,hertz2022prompt,epstein2024diffusion,bhunia2023person,zhou2022cross,ma2023waveipt}. Text-prompt-based methods~\cite{rombach2022high,ramesh2022hierarchical,saharia2022photorealistic} often struggle with precise control, limited by text-image pair quality and complexity of lengthy inputs. Attention control~\cite{hertz2022prompt,epstein2024diffusion} and latent editing~\cite{nie2023blessing,chefer2023attend} can somewhat enhance alignment of text-to-image generation. Personalized methods~\cite{ruiz2023dreambooth,hu2021lora} capture the common concept across a few images and reproduce that concept in generated images. Furthermore, some research~\cite{zhang2023adding,mou2023t2i} introduces adapters that input various conditions like skeleton images to steer semantics and spatial structures. Our work centers on controllable video generation with pre-trained adapters and latent editing.

\subsection{Video Editing and Predicting}
Video editing aims to modify given videos to meet specific requirements. Training-based methods~\cite{esser2023structure,chen2023control,liew2023magicedit} involve collecting and training on large-scale videos, incurring high costs. To circumvent this, train-free methods~\cite{wang2023zero,zhang2023controlvideo,qi2023fatezero} effectively extend existing text-to-image models into the video domain but may compromise on capabilities. One-shot methods~\cite{wu2023tune,zhao2023controlvideo,liu2023video,shin2023edit,chai2023stablevideo} offer a balance between computational cost and performance, requiring tuning on just a specific video instance. Our work also focuses on the one-shot setting but surpasses video editing, as it eliminates the need for corresponding original videos of input pose conditions. Video prediction~\cite{xing2023survey,voleti2022mcvd,gu2023seer} is intended to forecast a segment of a video based on other segments. In our context, when training and inference poses originate from the same individual, our task can be viewed as video prediction.

\subsection{Personalization Video Generation}
Text-to-video models~\cite{ho2022imagen,singer2022make,khachatryan2023text2video} create a specific subject video through text description. Image-to-video approaches~\cite{blattmann2023stable,girdhar2023emu,ni2023conditional} animate a still image to produce a dynamic variation about it. To harness the vast array of personalized image models in the community, certain techniques~\cite{guo2023animatediff} transfer these into video generation models by incorporating temporal modeling modules pre-trained on a large-scale video dataset. Concentrating on controllable human-centric generation, some human pose transfer works~\cite{zhao2022thin,siarohin2021motion} migrate the motions of a source video to a target subject. However, these methods are inflexible due to using the original video as guidance. More relevant to our work, pose-guided image-to-video generation~\cite{xu2023magicanimate,wang2023disco,hu2023animate,chang2023magicdance} adapts a given image to match specific pose sequences. Due to the complex distribution of human video data, they often need to train on extensive high-quality human videos that may be unavailable. To circumvent the need for large-scale human video datasets and extensive training resources, we adopt a one-shot setting. Disco~\cite{wang2023disco}, which underwent training on the TikTok dataset~\cite{jafarian2021learning} offers a strategy for fine-tuning on a single video. In contrast, our method eliminates the need for pre-training on video data, requiring tuning only on a single video of the target individual. Similar to our work, Everybody-Dance-Now~\cite{chan2019everybody} excludes additional training videos but requires long videos of a target subject (8-17 minutes), while our method merely needs a few video frames (e.g., 8 frames).

\section{Preliminary}
\label{sec:controlvideo}

ControlVideo~\cite{zhao2023controlvideo} is a representative one-shot video editing method following the additional control based on ControlNet~\cite{zhang2023adding}.
For video editing, formally, given a source prompt $p_s$, a reference video with $N$ frames $\mathbf{X} = \{\mathbf{x}_i\}_{i=1}^N$, a target prompt $p_t$, and extracted condition $ \mathbf{Q}$ from $\mathbf{X}$, it aims to generate a high-quality video $\mathbf{V} = \{\mathbf{v}_i\}_{i=1}^N$.
In the following, we present the technical details of ControlVideo.

\subsubsection{Model Architecture.} 
\label{sec:controlvideo_model}
ControlVideo consists of two main components: a large-scale latent diffusion model (i.e. Stable Diffusion) for image generation~\cite{rombach2022high}, and a condition control model, ControlNet~\cite{zhang2023adding}, which ensures the generated images adhere to input conditions $\mathbf{Q}$. To adapt this image generation framework to video applications, ControlVideo integrates two key temporal modules: key-frame attention and temporal attention.

\subsubsection{Key-Frame Attention.} 
\label{sec:controlvideo_time}
To achieve an intersection in the temporal axis across frames, ControlVideo advances self-attention as key-frame attention, in which all frames in a video share the key and value of a certain frame. Specifically, for a hidden feature map $x \in \mathbb{R}^{(f, d, c)}$ where $f$ indicates video length and $d$ is the 1D spatial dimension, given a query matrix $Q_i$, key matrix $K_i$, and value matrix $V_i$ in a self-attention for the $i$-th frame, key-frame attention is formulated as: 
\begin{align}
\label{eq:key-frame}
    \textrm{Softmax}(\frac{Q_i K_k^T}{\sqrt{c}}) V_k,
\end{align}
where $c$ is the hidden feature dimension and $k$ is the index of the key frame providing the key and value for all frames. 

\subsubsection{Temporal Attention.} To further improve temporal consistency, ControlVideo introduces temporal attention~\cite{ho2022video}, which computes feature similarity at the same spatial location across all frames. Specifically, it firstly exchanges the frame length and the spatial dimension so that $x \in \mathbb{R}^{(d, f, c)}$. Then, given the $Q_j$, $K_j$, and $V_j$ for the j-th spatial location, temporal attention is computed as:
\begin{align}
\label{eq:temporal}
    \textrm{Softmax}(\frac{Q_j K_j^T}{\sqrt{c}})V_j.
\end{align}

\subsubsection{Training and Inference.} 
\label{sec:training_and_indference}
To enable parameter-efficient fine-tuning based on pre-trained weights from large-scale image generation models~\cite{rombach2022high}, partial parameters in key-frame attention and temporal attention are trainable. In particular, ControlVideo model $\bm\epsilon_{\bm\theta}$ can be trained on a single training video with extracted conditions $\mathbf{Q}$ by the mean-squared error:

\begin{align}
\label{eq:training_loss}
\mathbb{E}_{t, \mathbf{X}, \bm\epsilon \sim \mathcal{N}(\mathbf{0},\textbf{I})}||\bm\epsilon - \bm\epsilon_{\bm\theta}(\mathcal{E}(\mathbf{X}_t), t, \mathbf{Q}, p_s)||^2_2,
\end{align}
where $\mathbf{X}_t$ indicates a noise video at the timestep $t$, and $\mathcal{E}(\cdot)$ is an encoder function compressing $\mathbf{X}_t$ into a lower-dimensional latent variables $\mathbf{Z}_t=\mathcal{E}(\mathbf{X}_t)$ which is reversible through a decoder function $\mathcal{D}(\cdot)$~\cite{rombach2022high}. During inference, it employs DDIM inversion ~\cite{song2020denoising} of $\mathbf{X}$ to create an initial latent $\mathbf{Z}_T$ as guidance, which is a crucial step for detail preservation of $\mathbf{X}$ and is represented as:
\begin{align}
\label{eq:contrlvideo_DDIM}
\mathbf{Z}_T = \textrm{DDIM-inv}(\mathbf{X}, \mathbf{Q}, p_s).
\end{align}

\begin{figure}[tb]
  \centering
  \includegraphics[width=1.0\linewidth]{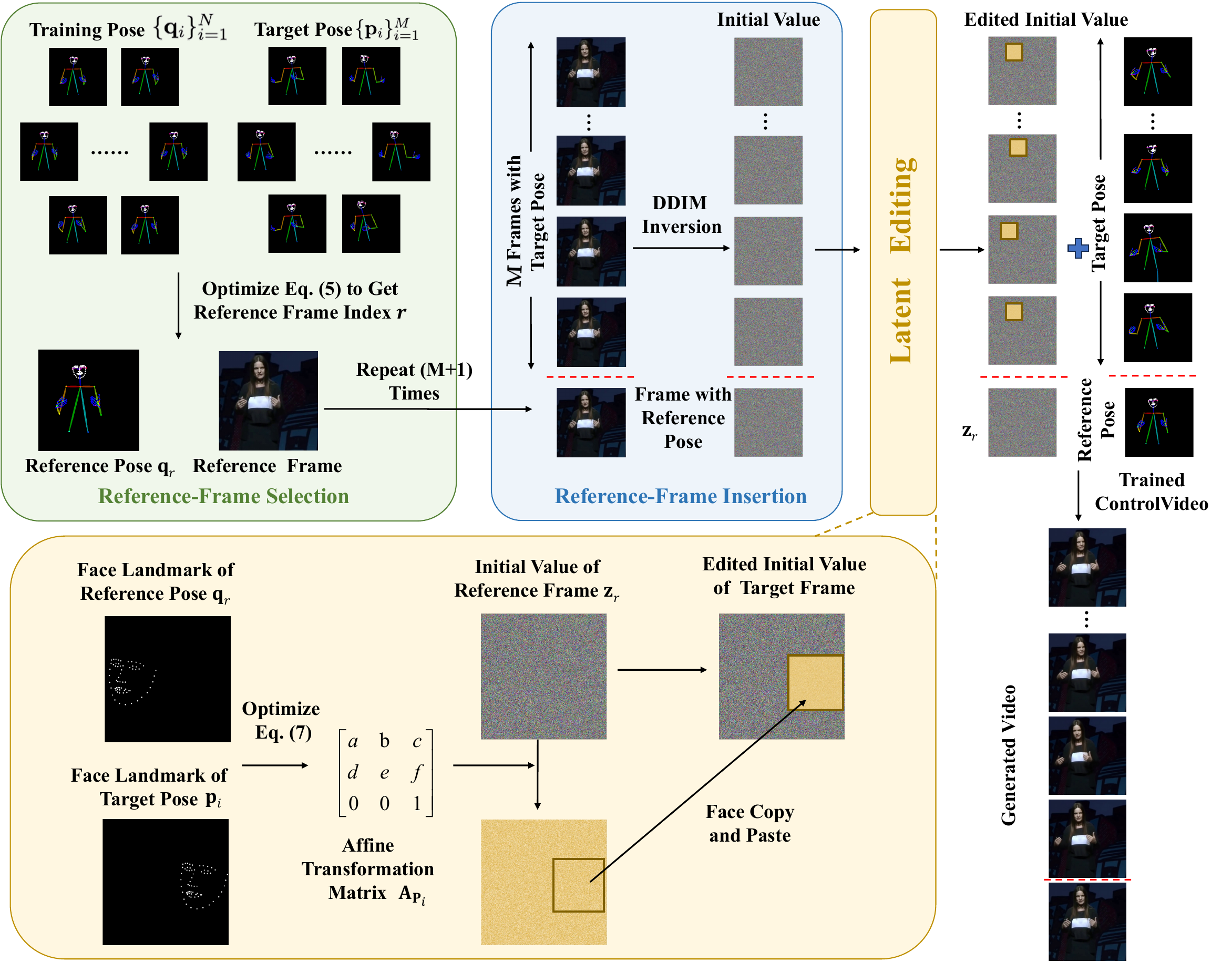}
  \caption{Inference framework of PoseCrafter. First, we select a frame from the training video to form a pseudo reference video, followed by DDIM inversion. Next, the pose from the reference frame is inserted into the inference poses. Finally, we edit latent to refine the generation of faces and hands by affine transformation.}
  \label{fig:method}
\vspace{-10pt} 
\end{figure}

\section{PoseCrafter}
\label{sec:method}

For crafting pose-guided personalized videos in a one-shot setting, we introduce a holistic framework PoseCrafter in \cref{sec:framework} and propose two key techniques, i.e. reference-frame selection and insertion in \cref{sec:ref-sel-ins} and latent editing in \cref{sec:latent-editing}.

\subsection{Settings and Framework}
\label{sec:framework}

Our goal with a human-centric reference video is to create a high-quality video that follows a specific target pose sequence while preserving the identity from the reference. This task is notably more complex than video editing due to the absence of corresponding videos for the target poses. Applying ControlVideo in this scenario is not straightforward, as it depends on the DDIM inversion of non-existent videos to generate the initial latent $\mathbf{Z}_T$ for guidance (see \cref{eq:contrlvideo_DDIM}).

Formally, given a source prompt $p_s$, a reference video with $N$ frames $\mathbf{X} = \{\mathbf{x}_i\}_{i=1}^N$ and corresponding extracted poses $\mathbf{Q} = \{\mathbf{q}_i\}_{i=1}^N$, along with a target prompt $p_t$ and a pose sequence consisting of $M$ frames $ \mathbf{P} =\{\mathbf{p}_i\}_{i=1}^{M}$, we aims to generate a high-quality video $\mathbf{V} = \{\mathbf{v}_i\}_{i=1}^M$. This video should be of superior fidelity, exhibit exceptional faithfulness to $\mathbf{X}$, and accurately align with $\mathbf{P}$.

As illustrated in ~\cref{fig:method}, we introduce a novel framework PoseCrafter, which shares a similar model and training protocol as ControlVideo yet a distinct inference process to address the aforementioned challenge. In particular, we utilize Stable Diffusion~\cite{rombach2022high} for video generation, coupled with the condition control model ControlNet~\cite{zhang2023adding} to accurately dictate specific postures in digital humans (see \cref{sec:controlvideo_model}). The training employs the loss function (see \cref{eq:training_loss}) on an individual video. Moreover, to mitigate GPU memory constraints on long video training, we randomly sample $n$ frames from a training video along the time dimension as a minimal batch at each training iteration. For the inference phase, to bypass the reliance on videos corresponding to target poses $\mathbf{P}$, we introduce a reference-frame selection and insertion technique to enhance faithfulness (see \cref{sec:ref-sel-ins}), and latent editing to refine face and hand details (see \cref{sec:latent-editing}).

\subsection{Reference-Frame Selection and Insertion}
\label{sec:ref-sel-ins}

Our preliminary experiment suggests that using random noise as the initial latent $\mathbf{Z}_T$ struggles to preserve human identity and background. To tackle this problem, we select the frame  $\mathbf{x}_{r}$ $(r \in [1, N])$ whose pose coordinates are closest to the inference poses from the training video as the reference frame. Given training poses $\{\mathbf{q}_i\}_{i=1}^N$ and inference poses $\{\mathbf{p}_i\}_{i=1}^M$, we determine the index of the reference frame $\mathbf{x}_{r}$ as follows:
\begin{align}
\label{eq:referecen_frame_selection}
r = \arg\min_{1 \leq i \leq N} \sum_{1 \leq j \leq M} \|\mathbf{q}_i - \mathbf{p}_j\|_2^2,
\end{align}
where $\|\cdot\|_2$ denotes the $\ell_2$ norm. 
Subsequently, we replicate $\mathbf{x}_{r}$ $M$ times to create a pseudo reference video $\{\mathbf{x}_{r}\}^M$, matching the length of the inference poses. This allows us to use the DDIM inversion~\cite{song2020denoising} of this pseudo video as the initial latent.

Notably, by inverting the reference video $\{\mathbf{x}_r\}^M$ to initial latent $\mathbf{Z}_T= \{\mathbf{z}_r\}^M$ and then generating the video $\mathbf{V}$ guided by inference poses $\mathbf{P}$, we can endeavor to preserve as much detail as possible in the training video. Previous work~\cite{wang2023disco} uses the same fixed noise latent for all generated frames rather than different random noise latent to maintain better temporal consistency, so we prefer merely one reference frame instead of multiple frames to form the pseudo reference video. 

To further enhance identity faithfulness, we insert the pose $\mathbf{q}_r$ of $\mathbf{x}_r$ into inference poses $\{\mathbf{P}_i\}_{i=1}^M$, creating an extended sequence of $M+1$ elements, i.e. $\{\mathbf{q}_r, \mathbf{p}_1,\mathbf{p}_2,\dots,\mathbf{p}_M\}$\footnote{To be compatible with key-frame attention, we place $\mathbf{q}_r$ as the key frame with position $k$, and in our experiments, we set $k=1$ by default. }. Following this augmented sequence, we generate $M+1$ frames and subsequently remove the redundant frame after inference. This method effectively deepens the connection between the generated video and the reference frame, i.e. faithfulness, as DDIM inversion and sampling \cite{song2020denoising} can accurately reconstruct the original image (i.e. the reference frame) while temporal modules (see \cref{sec:controlvideo_time}) encourage generated frames to be consistent.

\subsection{Improve Face and Hand Generation by Editing $\mathbf{Z}_T$}
\label{sec:latent-editing}

Our initial experiments reveal that a higher number of landmarks for faces and hands, coupled with the demand for finer details in the generated results, necessitates a tailored approach to editing latent variables for facial and hand features to enhance generation quality. Specifically, we draw inspiration from classical methods in face alignment~\cite{jabberi202368}, calculating the corresponding affine matrix based on facial landmarks from the reference frame pose and the target frame pose.
More precisely, given a face landmark $(x,y)$ and its corresponding target $(x', y')$, we can achieve that mapping by an affine transformation:
\begin{align}
\label{eq:affine_transformation}
\left[\begin{array}{c}
x' \\
y' \\
1
\end{array}\right] = \mathbf{A} 
\left[\begin{array}{c}
x \\
y \\
1
\end{array}\right] = \left[\begin{array}{ccc}
a & b & c \\
d & e & f \\
0 & 0 & 1
\end{array}\right] 
\left[\begin{array}{c}
x \\
y \\
1
\end{array}\right],
\end{align}
where $\mathbf{A}$ is an affine transformation matrix with six degrees of freedom. 
For scenarios involving multiple landmarks, we solve the following least square optimization problem to get the affine transformation matrix $\mathbf{A}_{\mathbf{p}_{i}}$:
\begin{align}
  \mathbf{A}_{\mathbf{p}_{i}} =  \arg\min_{\mathbf{A}} \sum_{\mathbf{v} \in F_{\mathbf{q}_r}} \|\mathbf{A} \mathbf{v}  - \mathbf{v}_{\mathbf{p}_{i}}\|_2^2,
\end{align}
where $F_{\mathbf{q}_r}$ consists of all face landmarks for the reference pose $\mathbf{p}_r$ and $\mathbf{v}_{\mathbf{p}_{i}}$ is the landmark in the target pose $\mathbf{p}_{i}$ corresponding to $\mathbf{v}$.

As illustrated in \cref{fig:method}, for each inference pose $\mathbf{p}_i$, we edit its latent variable $\mathbf{z}_r$, derived from the DDIM inversion process, using the specific affine matrix $\mathbf{A}_{\mathbf{p}_{i}}$. Specifically, we apply the affine transformation to the entire latent variable $\mathbf{z}_r$, creating a transformed version $\mathbf{z}_r^{\mathbf{p}_i}$. We then selectively transfer the facial area of $\mathbf{z}_r^{\mathbf{p}_i}$, defined as the minimal bounding rectangle encompassing the facial landmarks of $\mathbf{p}_i$, directly onto the corresponding areas of $\mathbf{z}_r$ for generating target frame $\mathbf{v}_i$. Moreover, similar operations based on hand landmarks are also performed, resulting in improved hand features in our experiments (see \cref{fig:qualitive_ablation}). We believe that this latent editing can be extended to more body parts like legs.

It is noted that the latent variable for each frame is a copy of the latent variable from a single reference frame, hence we perform the DDIM inversion operation just once. Subsequently, different affine transformation matrices are applied to obtain corresponding final latent variables. This approach is more efficient than performing affine transformations on the reference frame followed by separate DDIM inversions for each. Moreover, editing directly in pixel space reduces image quality obtained after DDIM inversion and subsequent sampling.

Notably, PoseCrafter can follow poses from different individuals or artificial edits and simultaneously retain the identity in an open-domain training video.

\section{Experiments}
\label{sec:experiments}

\subsection{Experimental Settings}

\subsubsection{Datasets.} 
We train and evaluate our method on test splits of two public datasets TED~\cite{siarohin2021motion} and TikTok~\cite{jafarian2021learning}. Due to changes in user privacy and download permissions, we can only download 128 TED test videos from YouTube using download links provided in~\cite{siarohin2021motion}. Following prior work~\cite{wang2023disco,xu2023magicanimate,chang2023magicdance}, we use 10 TikTok test videos presented in~\cite{wang2023disco}. We also gather 10 high-quality videos from YouTube, encompassing a variety of scenes such as interviews, movie clips, and talk shows. For each video, we designate 100 frames as test frames and uniformly select $N$ frames from their preceding frames as training frames. We crop and resize all videos to a uniform $512\times512$ resolution. See Appendix A.1 for more details.

\vspace{-10pt} 
\subsubsection{Implementation Details.} We initialize the models with the publicly available realistic model, leosamsMoonfilm~\cite{CivitaiModel2023}, based on Stable Diffusion 1.5. For accurate pose extraction, we employ DWpose~\cite{yang2023effective}, an excellent 2D pose extractor. During training, we set the learning rate as 0.003 and the minimal batch size $n=8$. Note that our method requires only 2 minutes to complete training on an individual video with 8 frames using an RTX3090 GPU, and all experiments are implemented on RTX3090 GPUs. Please refer to Appendix A.2 for more details.

\subsubsection{Evaluation.} 

We evaluate video quality encompassing temporal consistency, faithfulness, fidelity, and pose alignment by 8 common quantitative metrics. Specifically, we measure temporal consistency by calculating the average CLIP similarity (CLIP-T) among all frames of the generated video. To assess faithfulness, we use the average similarity (CLIP-I) between reference images and generated frames. Regarding video fidelity, we adopt FID and FVD. For pose alignment, we compute the mean squared error (MSE-P) between 2D keypoint coordinates of all pose pairs extracted from generated images and real images. Additionally, we consider common reconstitution measures such as SSIM, PSNR, and LPIPS, which calculate the average distance between all pairs of generated images and corresponding ground truth, involving temporal consistency, faithfulness, fidelity, and pose alignment.

\vspace{-10pt}
\subsubsection{Baselines.} We select four representative methods as baselines for a comprehensive comparison. (1) Pose-guided image-to-video methods: Disco~\cite{wang2023disco} and MagicAnimate~\cite{xu2023magicanimate}, which are pre-trained on a combination of large-scale image data and smaller video data. We utilize their provided public checkpoints trained on TikTok videos (around 350 videos with 10-12 seconds and 30 fps)\footnote{Disco additionally collects 250 internal TikTok-style videos for training.}. (2) Pose-guided one-shot fine-tuning methods: ControlVideo and fine-tuned Disco. For ControlVideo, we inverse the 1-st frame in the training video and edit it guided by inference poses. For Disco, we use their introduced methods to fine-tune it on each video. More details are provided in Appendix A.3.

\subsection{Main Results}
\label{sec:main_results}
We demonstrate that PoseCrafter fine-tuned on a single video achieves superior quantitative and qualitative results to baselines pre-trained on a vast collection of videos. Note that, for a fair comparison, we unify image resolution, test dataset, and evaluation codes which may differ across all methods.

\vspace{-10pt} 
\subsubsection{Quantitative Comparisons.} \cref{tab:quantitative_results_TikTok} and \cref{tab:quantitative_results} present the quantitative comparison on TikTok and TED, separately. Although training on 8 frames, by the comprehensive framework and sophisticated inference algorithm, PoseCrafter achieves competitive results on all 8 metrics to the state-of-the-art baseline MagicAnimate on TikTok, but superior results than it on TED which indicates its limited generalization capability. Moreover, our method significantly improves all metrics than ControlVideo and one-shot Disco, demonstrating the superiority of our training and inference framework.  Remarkably, PoseCrafter, which requires only about 12 minutes of training on 32 frames with a single RTX3090 GPU, significantly outperforms MagicAnimate and excels in all quantitative metrics than all baselines on TikTok. This underscores the potential of one-shot methods, which, even with limited training, can surpass approaches relying on extensive training for crafting high-quality pose-guided personalized videos.

Additionally, fine-tuned Disco degrades many metrics on TikTok. We hypothesize that Disco fine-tunes its cross-attention layers in UNet and full parameters in ControlNet (294.95 M in total), and thus is easy to overfit to a single training video. As a comparison, our method only requires fine-tuning 68.79M trainable parameters (see \cref{sec:training_and_indference}), which is a more efficient fine-tuning strategy. Moreover, given the variations in evaluation codes, generated image resolutions and test frame counts among methods, we avoid directly citing their reported results. Instead, we standardize these elements to ensure a fair comparison. A comparison with their originally reported results is provided in Appendix B.1.

\begin{table*}[tb]
\centering
\caption{
Quantitative results on TikTok test dataset. NFs represent the number of available frames for the corresponding method. Image-GT signifies that these image-level metrics are calculated between each generated image and its corresponding ground truth image. $\dagger$ represents we fine-tune corresponding methods using their tailored strategies for specific subjects.
}
\label{tab:quantitative_results_TikTok}
\resizebox{\textwidth}{!}{
\begin{tabular}{lccccccccc} 
\toprule
\multirow{2}[3]{*}{Method} & \multirow{2}[3]{*}{NFs} & \multicolumn{4}{c}{\textbf{Image-GT}} & \multicolumn{2}{c}{\textbf{Image}} & \multicolumn{2}{c}{\textbf{Video}} \\ 
\cmidrule(lr){3-6} \cmidrule(lr){7-8} \cmidrule(lr){9-10}
 & & \textbf{SSIM}\ $\uparrow$ & \textbf{PSNR}\ $\uparrow$ & \textbf{LPIPS}\ $\downarrow$ & \textbf{MSE-P}\ $\downarrow$ & \textbf{FID}\ $\downarrow$& \textbf{CLIP-I}\ $\uparrow$ & \textbf{FVD}\  $\downarrow$ & \textbf{CLIP-T} $\uparrow$ \\

\midrule
DisCo~\cite{wang2023disco}  & 1 & 0.704
& 15.16 & 0.358 & 9.11E-3
 & 76.40 & 0.827 & 689.23 & 0.908\\
MagicAnimate~\cite{xu2023magicanimate} & 1
 &  0.756 & 17.95 & 0.265 & 5.48E-3  & 57.60 & 0.846 & 374.40& 0.918\\ 
\midrule
Disco$\dagger$~\cite{wang2023disco} & 8 & 0.683 & 15.33 & 0.371 & 8.86E-3 & 65.73 & 0.813 & 807.35& 0.886 \\ 
ControlVideo~\cite{zhao2023controlvideo} & 8 & 	0.738 & 16.93 & 0.311 & 6.78E-3 & 56.50 & 0.827 & 489.30 &	0.912
  \\ 
PoseCrafter (\textbf{ours}) & 8
& 0.765
 & 17.36 & 0.275 & 5.76E-3 & 48.09
& 0.840 & 440.49& 0.921\\
\midrule
PoseCrafter (\textbf{ours}) & 16 &  0.776 & 17.87 & 0.252 & 5.23E-3 & 42.09
& \textbf{0.864} &	397.19 & 0.919 \\

PoseCrafter (\textbf{ours}) & 32 & \textbf{0.786}
 & \textbf{18.56} & \textbf{0.233}
 & \textbf{4.05E-3} & \textbf{39.65}
& 0.854 & \textbf{362.09} & \textbf{0.922} \\
\bottomrule
\end{tabular}
}
\end{table*}

\begin{table*}[tb]
\centering
\caption{Quantitative results on TED test dataset. 
NFs represent the number of available frames for the corresponding method. Image-GT signifies that these image-level metrics are calculated between each generated image and its corresponding ground truth image. $\dagger$ represents we fine-tune corresponding methods using their tailored strategies for specific subjects.
}
\label{tab:quantitative_results}
\resizebox{\textwidth}{!}{
\begin{tabular}{lccccccccc}
\toprule
\multirow{2}[3]{*}{Method} & \multirow{2}[3]{*}{NFs} & \multicolumn{4}{c}{\textbf{Image-GT}} & \multicolumn{2}{c}{\textbf{Image}} & \multicolumn{2}{c}{\textbf{Video}} \\ 
\cmidrule(lr){3-6} \cmidrule(lr){7-8} \cmidrule(lr){9-10}
 & & \textbf{SSIM}\ $\uparrow$ & \textbf{PSNR}\ $\uparrow$ & \textbf{LPIPS}\ $\downarrow$ & \textbf{MSE-P}\ $\downarrow$ & \textbf{FID}\ $\downarrow$& \textbf{CLIP-I}\ $\uparrow$ & \textbf{FVD}\  $\downarrow$ & \textbf{CLIP-T} $\uparrow$ \\

\midrule
DisCo~\cite{wang2023disco}  & 1 &  0.550 & 17.22 & 0.327 & 3.20E-3& 51.57& 0.794 & 309.62 & 0.909
\\
MagicAnimate~\cite{xu2023magicanimate} & 1
 &  0.498 & 13.32 & 0.338& 1.97E-3 & 46.96 & 0.801 & 194.03& 0.912
\\ 
\midrule
Disco$\dagger$~\cite{wang2023disco} & 8 & 0.551 & 17.98 &  0.373 & 3.93E-3 & 60.56 & 0.827 & 244.62 & 0.909\\ 
ControlVideo~\cite{zhao2023controlvideo} & 8 &  0.771  & 22.32 & 0.196 & 1.62E-3 &29.01 & 0.866 & 109.36 & 0.940\\ 
PoseCrafter (\textbf{ours}) & 8
& \textbf{0.810} & \textbf{23.92} & \textbf{0.142} & \textbf{1.56E-3} & \textbf{20.40}
& \textbf{0.906} & \textbf{80.01} & \textbf{0.954}\\

\bottomrule
\end{tabular}
}
\vspace{-10pt}
\end{table*}

\vspace{-10pt} 
\subsubsection{Qualitative Comparisons.} 
\begin{figure}[htp]
  \centering
  \includegraphics[width=1.0\linewidth]{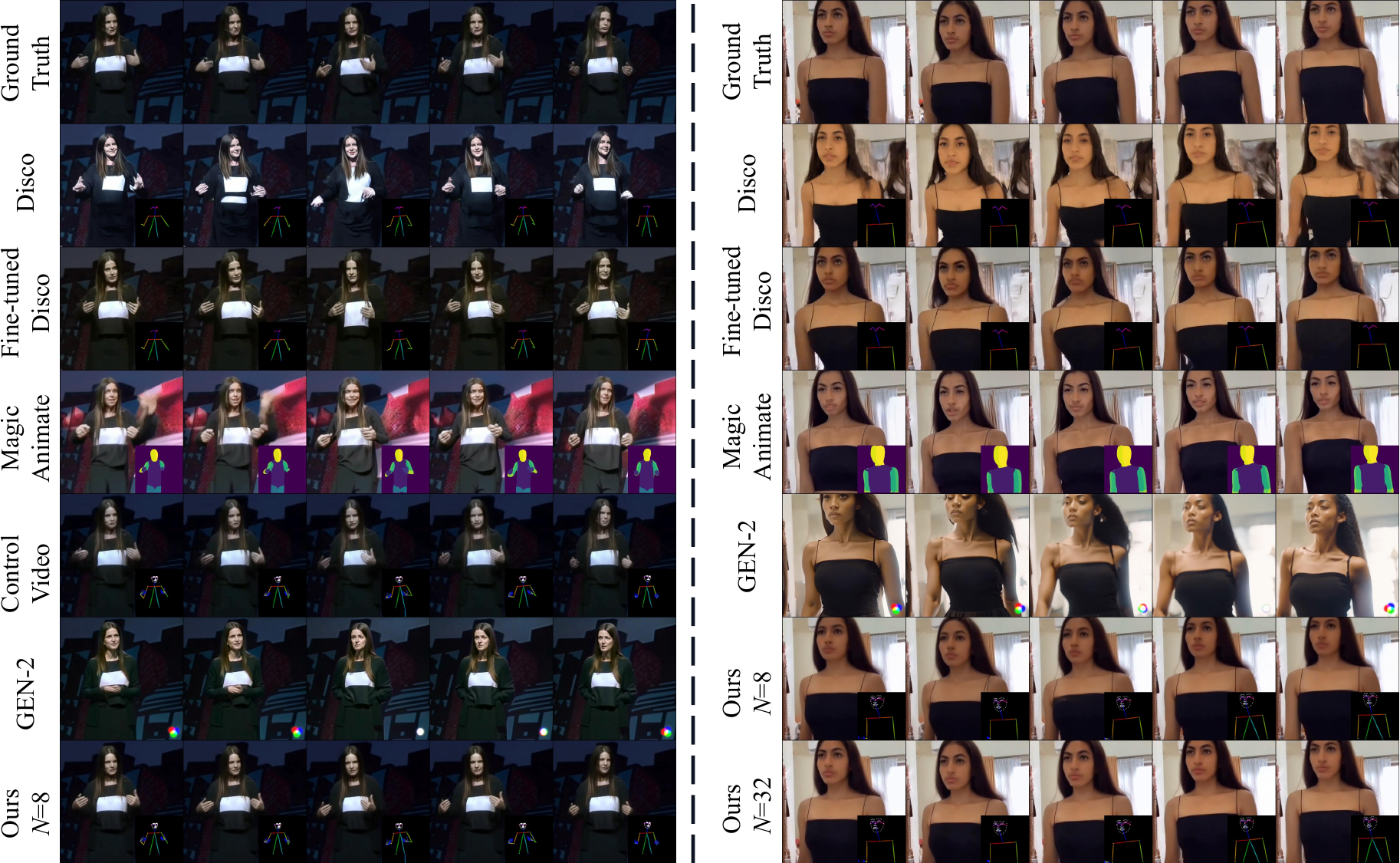}
  \caption{Qualitative Comparisons of all methods ($M=100$). Left: TED. Right: TikTok. Time progresses from left to right. PoseCrafter yields the highest quality videos.}
  \label{fig:qualitive_comparation}
\vspace{-10pt}
\end{figure}

\cref{fig:qualitive_comparation} shows the qualitative comparisons of our method to all baselines on TED and TikTok. Disco and MagicAnimate rely on OpenPose~\cite{cao2017realtime} or DensePose~\cite{guler2018densepose} without hand and face poses and thus do not precisely control hand and face details such as facial expression. Moreover, due to the absence of similar reference videos in their training corpus, they struggle to preserve high-level human details such as illumination and clothing on TED. Although fine-tuned Disco on each individual improves detail maintenance to training videos and ControlVidoe uses the DDIM inversion, they still do not supersede PoseCrafter, emphasizing the importance of our training and inference framework. As time progresses, the characters generated by a commercial-grade application ( i.e. GEN-2) increasingly diverge from the reference image. 

Additionally, with more training frames ($N=32$), PoseCrafter generates the highest quality videos among all methods on TikTok, which is consistent with quantitative results. Overall, PoseCrafter yields videos most closely aligning with the ground truth among all approaches on TED and TikTok. Remarkably, when the extracted poses are precise, PoseCrafter is capable of generating videos of superior quality compared to the original ground truth, exemplified by our hands in the last image of the TED examples in \cref{fig:qualitive_comparation}.

\begin{figure}[htp]
  \centering
  \includegraphics[width=1.14 \linewidth]{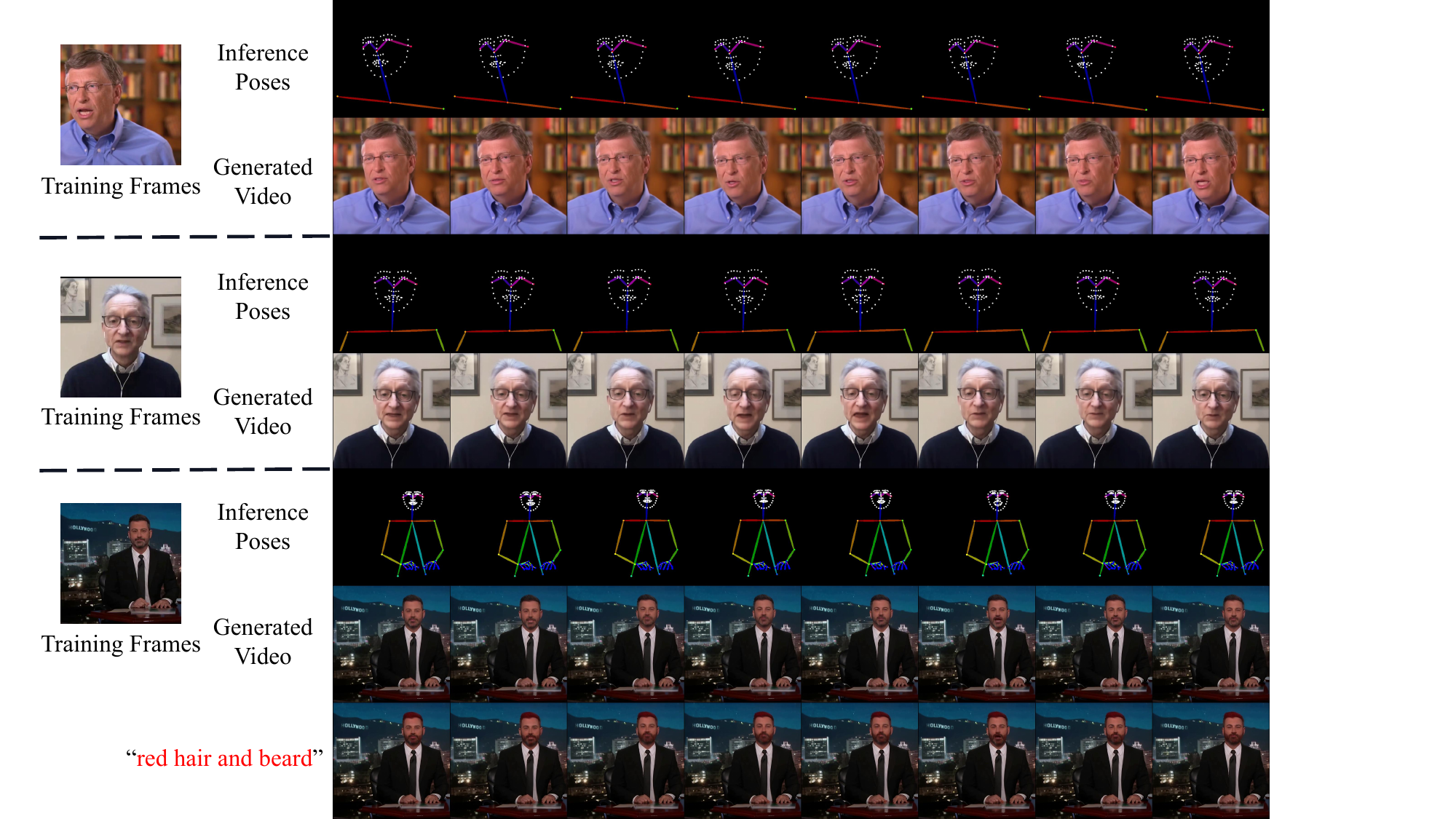}
  \caption{Inference from poses of the same individual ($N=100$ and $M=100$). Time progresses from left to right. }
  \label{fig:inference_poses}
\vspace{-10pt} 
\end{figure}

\begin{figure}[htb]
  \centering
  \includegraphics[width=1.0\linewidth]{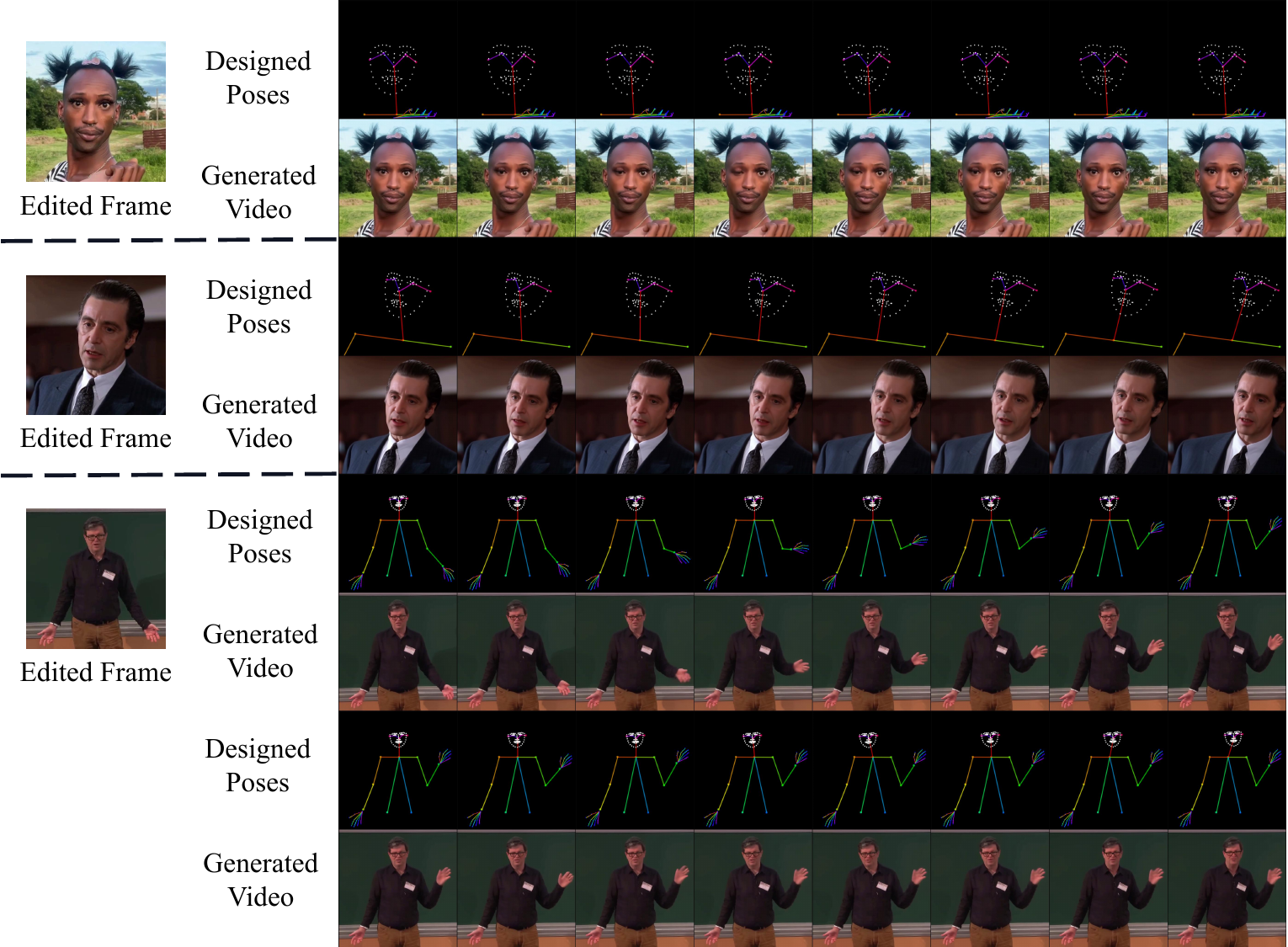}
  \caption{Inference from artificially designed poses ($N=100$ and $M=8,8,16$). Time progresses from left to right. From top to bottom: the first pose 
  sequence features a right eye blink; the second pose sequence involves tilting the head rightward; the third pose sequence is waving an arm followed by shaking the head.}
  \label{fig:inference_designed_poses}
\vspace{-10pt} 
\end{figure}

\begin{figure}[htb]
  \centering
  \includegraphics[width=1.0\linewidth]{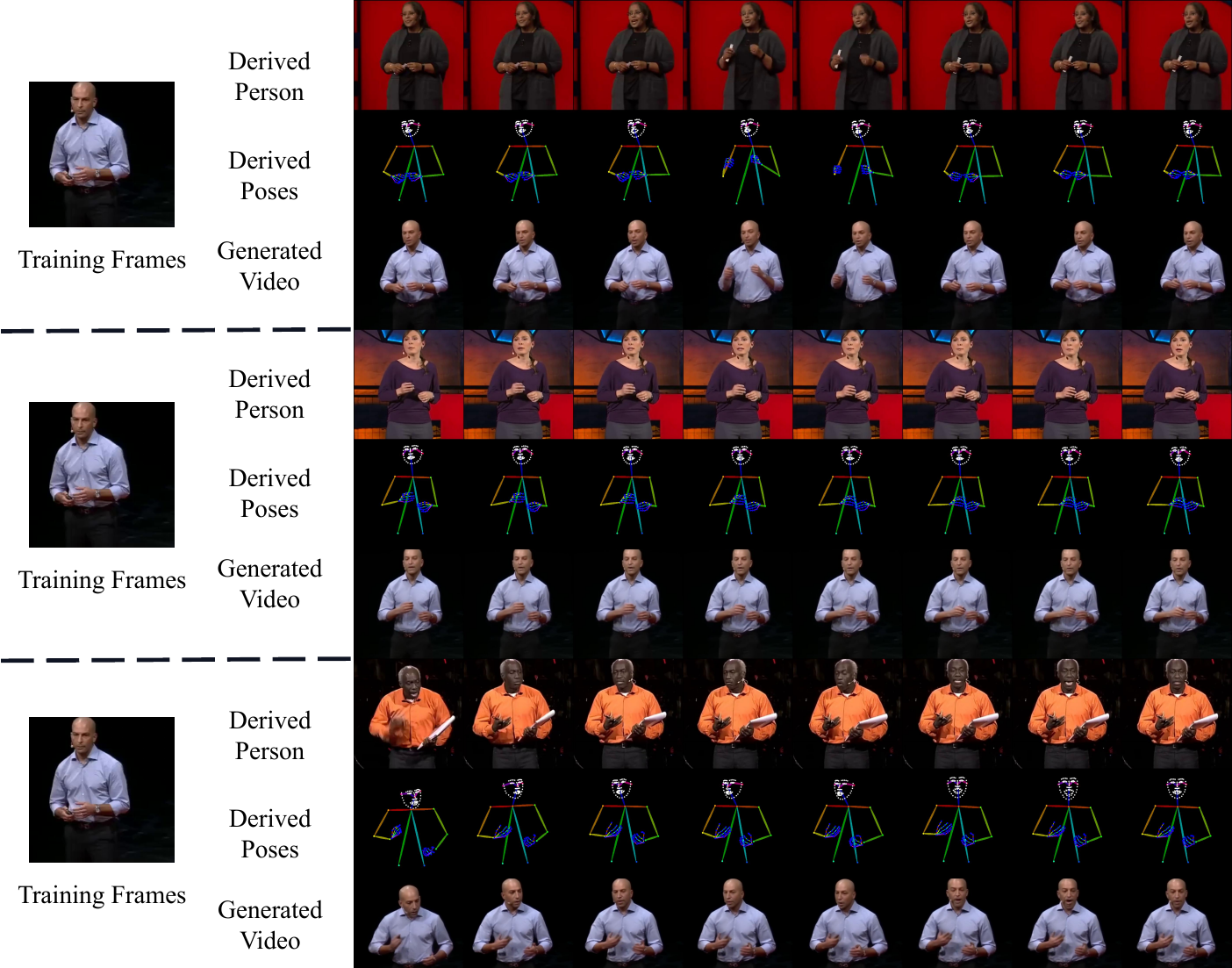}
  \caption{Inference poses of other individuals ($N=50$ and $M=50$). Time progresses from left to right. } 
  \label{fig:Inference_other_person_poses}
\vspace{-10pt} 
\end{figure}

\vspace{-10pt} 
\begin{figure}[htb]
  \centering
  \includegraphics[width=0.8\linewidth]{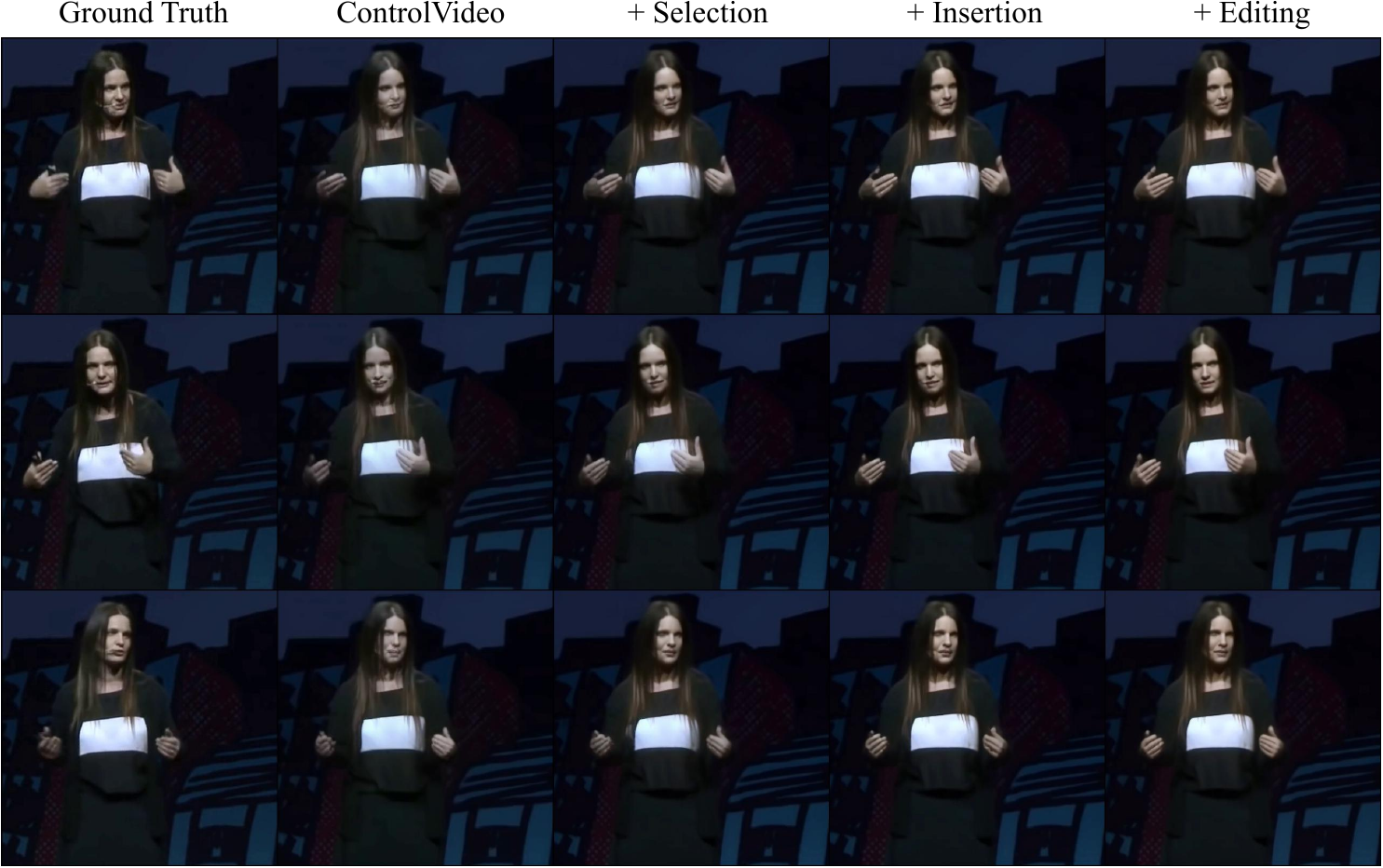}
  \caption{Qualitative ablation of PoseCrafter.}
  \label{fig:qualitive_ablation}
\vspace{-10pt}
\end{figure}

\vspace{-10pt}
\subsection{Applications}
\label{sec:applications}
PoseCrafter, trained in an open-domain video such as interviews, movie clips, and speeches, can follow flexible pose control: (1) poses from the same individual; (2) artificially crafted poses; and (3) poses from different individuals.

\vspace{-10pt}
\subsubsection{Inference with Poses from the Same Individual.} 
\cref{fig:inference_poses} depicts the results of PoseCrafter guided by poses of the same individual as in training frames. PoseCrafter adeptly reconstructs intricate background elements like picture frames and logo texts, and human details including hairstyles and wrinkles. By tweaking the target prompt $p_t$, it allows for modifications in human attributes, such as changing hair color from black to red, as illustrated in the final row of \cref{fig:inference_poses}. In cases where partial segments in a video are of low quality yet yield reliable poses, we can utilize PoseCrafter trained on remaining segments to regenerate the deficient clips, serving as a form of video inpainting.

\vspace{-10pt}
\subsubsection{Inference with Artificially Designed Poses.} 
Using pose editors~\cite{huchenlei_sdwebuiopenposeeditor_2023}, we can alter the pose of a specific training frame to create ``dream pose sequences'' for guiding PoseCrafter. Illustrated in \cref{{fig:inference_designed_poses}}, we demonstrate this by having a character in entertainment videos progressively blink their right eye, a movie character gradually turn their head to the right, and a speaker lift and wave their left hand and finally sway their head.

\vspace{-10pt}
\subsubsection{Inference with Poses from Different Individuals.} 
As depicted in \cref{fig:Inference_other_person_poses}, PoseCrafter effectively utilizes the poses of an individual different from that in the training frames, successfully creating consistent and faithful videos. However, there may be variations in the facial and body shapes of the created characters, owing to the use of poses from figures with varying facial and body proportions. Moreover, more results with larger and more complex movements are provided in Appendix B.2.

\vspace{-5pt}
\subsection{Ablation}

We investigate the significance of reference-frame selection, integration, and latent editing. As demonstrated in the \cref{fig:qualitive_ablation}, the automatic selection of a reference frame in original videos leads to a notable improvement in video quality and faithfulness compared to ControlVideo. Its insertion into inference poses further enhances detail preservation in training videos. In addition to refining face and hand generation, latent editing strengthens temporal consistency in generated videos. The quantitative analysis, detailed in Appendix C.1, corroborates the effectiveness of these distinct stages in our methodology. In summary, these critical design elements contribute to the enhancement of video quality. Furthermore, the roles of training frame quantity and sample timestep choice in latent editing are explored in Appendices C.2 and C.3, respectively.

\vspace{-5pt}
\section{Conclusions}
\label{sec:conclusion}
In this paper, we introduce PoseCrafter, a one-shot tuning method for crafting pose-guided personalized videos without the need for extensive data. To address dependence on original videos of given pose sequences in video editing, we introduce two crucial algorithms to enhance video quality: reference frame selection and insertion, and latent editing. Furthermore, PoseCrafter, trained in an open-domain video, demonstrates adaptability to a range of pose sources, including poses from distinct individuals.

\vspace{-10pt}
\subsubsection{Limitations.} The video quality of PoseCrafter depends on the ControlNet and latent diffusion models, which struggle with complex poses like crossed fingers, presenting a limitation in our approach. When using poses from others, the body proportions of our generated characters change accordingly. Moreover, large differences between training and inference poses can degrade video quality. Specific failure cases are presented in Appendix B.4.

\clearpage  

\section*{Acknowledgements}

This work was supported by NSF of China (No. 62076145), Major Innovation \& Planning Interdisciplinary Platform for the ``Double-First Class" Initiative, Renmin University of China, the Fundamental Research Funds for the Central Universities, the Research Funds of Renmin University of China (No. 22XNKJ13), and the Huawei Group Research Fund. C. Li was also sponsored by Beijing Nova Program (No. 20220484044). The work was partially done at the Engineering Research Center of Next-Generation IntelligentSearch and Recommendation, Ministry of Education.

%
%
\bibliographystyle{splncs04}
\bibliography{main}

\clearpage

\appendix

\begin{center}
\textbf{\Large Supplementary Material}
\end{center}

\section{Experimental Settings}

In this section, we present more experimental details about datasets in \cref{sec:more_datasets}, implementation details in \cref{sec:more_implementation}, and baselines in \cref{sec:more_baseline_details}. We also detail the inference cost of PoseCrafter in \cref{sec:inference_cost} and discuss its potential negative impact in \cref{sec:potential_negative_impact}.

\subsection{Datasets}
\label{sec:more_datasets}

In order to fully, reasonably, and comprehensively evaluate methods, we collect 10 high-quality videos in the open domain from YouTube, encompassing a variety of scenes such as interviews, movie clips, and talk shows. The collected videos were already publicly available online. We emphasize that our intention is purely academic and no offensive edits or alterations have been made to the original content. Additionally, we extend our sincere acknowledgments and respect to the original video producers.

For each video on TED, we take 8 frames uniformly from the first 36 frames for training. For inference, we extract pose information of the consecutive 100 frames, beginning with the 46-th frame, thereby setting $N=8$ and $M=100$. For each video on TikTok and our collected dataset, due to the need to vary the number of training frames, we designate the final 100 frames as test frames (i.e. $M=100$) and uniformly select $N$ frames from their preceding frames as training frames. 

\subsection{Implementation Details}
\label{sec:more_implementation}

We initialize Pose ControlNet\footnote{\url{https://huggingface.co/lllyasviel/ControlNet-v1-1}} and VAE\footnote{\url{https://huggingface.co/stabilityai/sd-vae-ft-mse}} using their provided public checkpoints, without any additional fine-tuning on other data. The guidance scale of free-classifier guidance is set to 1 and increased to 3 for attribution editing. It is important to note that a larger guidance scale results in the generated video being more aligned with the target prompt, but potentially less faithful to the training video. We use the default control scale of ControlNet, which is set at 1. Additionally, we adopt the 50 DDIM sampling steps for inference.

We set the source prompt $p_s$ and target prompt $p_t$ as ``a person is speaking'' for TED videos and ``a person is dancing'' for TikTok videos. For other datasets, we use the default prompt ``a person''. For attribute editing, we append relevant text to the source prompt. For example, to modify the hair color of the generated character to red, we use the prompt ``a person, red hair'' corresponding to its source prompt ``a person''. 

We use the fixed learning rate of 0.003 and a fixed minimal batch size of 8 for all experiments. We set the max training step as 100 for 8 training frames and 2000 for 100 training frames. For the number of training frames $N$ other than 8 or 100, we calculate the max training step using the following formula:
\begin{align}
\label{eq:lr}
\text{round}(\frac{2000 - 100}{100 - 8} N + 100 - \frac{2000 - 100}{100 - 8}8) = \text{round}(\frac{475}{23}N - \frac{1500}{23}),
\end{align}
where $\text{round}(\cdot)$ denotes the function that rounds a value to the nearest integer.

It is worth noting that the aforementioned hyperparameters may not represent the optimal settings, but we empirically find that they can yield good results as defaults.

\subsection{Baselines}
\label{sec:more_baseline_details}

We use only the commercial application GEN-2 for qualitative comparison, adopting its default parameters in all experiments.  We utilize the prompt ``a person is speaking'' on TED and ``a person is dancing'' on TikTok.

We find that the default learning rate 1e-3 of Disco for human-specific fine-tuning tends to lead to overfitting, thus we adjust it to 1e-4 which yields better results. On TED, we use the default guidance scale of 3 for Disco but 1.5 for fine-tuned Disco to achieve better outcomes. On TikTok, we employ an optimal scale of 1.5, reported in~\cite{wang2023disco}, for both Disco and fine-tuned Disco. 

Regarding the selection of the reference image for image-to-video methods, we employ every frame from 8 training frames as a reference image for inference on TED, and report average quantitative results. On TikTok, since the training frames vary but share the same first frame, we designate this first frame as the reference image. Moreover, applying image-to-video methods across all training frames incurs substantial budgetary and time expenses, notably when $N=32$.

\subsection{Inference Cost} 
\label{sec:inference_cost}
PoseCrafter, with 1.48 billion parameters and around $1.747\times10^{14}$ FLOPs in total, takes 2.75 GPU minutes to generate 100 frames, utilizing 19.28 GB of memory on a single RTX 3090. In our experiments, using a single RTX 3090 with 24 GB of memory, we successfully generate videos up to a maximum length of 180 frames with good quality. PoseCrafter can generate longer videos with GPUs of larger memory, memory reduction techniques, and long-range video generation strategies.

\subsection{Potential Negative Impact} 
\label{sec:potential_negative_impact}
A major concern in human video generation is the risk of creating hyper-realistic videos that may impersonate real individuals. This technology allows for the production of avatars that closely resemble real people, often without their consent. Such convincing ``DeepFakes'' spark fears of identity theft, fraud, reputational harm, and regulatory challenges.

\section{Additional Results}
In this section, we present more related baselines in~\cref{sec:more_baselines}, more qualitative results in~\cref{sec:more_results}, more quantitative results in~\cref{sec:more_dataset_results}, and failure cases of PoseCrafter in~\cref{sec:failure_cases}.

\subsection{More Baselines}
\label{sec:more_baselines}
\cref{tab:app_quantitative_results_TikTok} and \cref{tab:app_quantitative_results} introduce additional image-to-video baselines, which are marked in gray, for TikTok and TED, respectively. TPS~\cite{zhao2022thin} and MRAA~\cite{siarohin2021motion} represents the state-of-the-art GAN-based methods and rely on ground truth videos. Consequently, \cite{xu2023magicanimate} develops an alternative version that requires only DensePose sequences. Moreover, \cite{xu2023magicanimate} establishes a related baseline, PA+CtrlN-V, integrating existing state-of-the-art image and video generation models. This includes the image-to-image method IP-Adapter~\cite{ye2023ip}, which maintains the identity of the reference image, Pose ControlNet~\cite{zhang2023adding}, controlling the pose in generated videos, and the text-to-video model AnimateDiff~\cite{guo2023animatediff}, ensuring temporal consistency in results. We also report the original quantitative results of Disco and MagicAnimate. 

It should be noted that these additional baselines cannot be directly compared with our method due to different settings in image resolution, the number of test frames, and evaluation codes. The original Disco trains and evaluates on 256-pixel resolution images, whereas we focus on a higher resolution of 512 pixels. Furthermore, while these image-to-video baselines evaluate using the full frames of test videos excluding reference images on TikTok and TED, we limit our testing to 100 frames for each video, as we require preceding frames for training. Regarding evaluation, MagicAnimate does not provide its implementation codes, and the Disco implementation contains errors\footnote{\url{https://github.com/Wangt-CN/DisCo/issues/86}}, especially for PSNR. Consequently, we have standardized these elements to ensure a fair comparison in the main text and detail them in the appendix.

\begin{table*}[tb]
\centering
\caption{
Quantitative results on TikTok test dataset. NFs represent the number of available frames for the corresponding method. Image-GT signifies that these image-level metrics are calculated between each generated image and its corresponding ground truth image. $\dagger$ represents we fine-tune corresponding methods using their tailored strategies for specific subjects. $\star$ indicates the results are directly cited from the \cite{xu2023magicanimate}. $\ddagger$ represents that the results are cited directly from the \cite{wang2023disco} where it additionally collects 250 internal TikTok-style videos for training.
}
\label{tab:app_quantitative_results_TikTok}
\resizebox{\textwidth}{!}{
\begin{tabular}{lccccccccc}
\toprule
\multirow{2}[3]{*}{Method} & \multirow{2}[3]{*}{NFs} & \multicolumn{4}{c}{\textbf{Image-GT}} & \multicolumn{2}{c}{\textbf{Image}} & \multicolumn{2}{c}{\textbf{Video}} \\ 
\cmidrule(lr){3-6} \cmidrule(lr){7-8} \cmidrule(lr){9-10}
 & & \textbf{SSIM}\ $\uparrow$ & \textbf{PSNR}\ $\uparrow$ & \textbf{LPIPS}\ $\downarrow$ & \textbf{MSE-P}\ $\downarrow$ & \textbf{FID}\ $\downarrow$& \textbf{CLIP-I}\ $\uparrow$ & \textbf{FVD}\  $\downarrow$ & \textbf{CLIP-T} $\uparrow$ \\

\midrule
\color[HTML]{9b9b9b}TPS$^\star$~\cite{zhao2022thin} & \color[HTML]{9b9b9b}1&  \color[HTML]{9b9b9b} 0.560 &  \color[HTML]{9b9b9b} 28.17 & \color[HTML]{9b9b9b}0.449& &  \color[HTML]{9b9b9b} 140.37 & &  \color[HTML]{9b9b9b}800.77& \\
\color[HTML]{9b9b9b}MRAA$^\star$~\cite{siarohin2021motion} & \color[HTML]{9b9b9b}1&  \color[HTML]{9b9b9b}0.646 &  \color[HTML]{9b9b9b}28.39 & \color[HTML]{9b9b9b}0.337& &  \color[HTML]{9b9b9b}85.49 & &  \color[HTML]{9b9b9b}468.66& \\
\color[HTML]{9b9b9b}IPA~\cite{ye2023ip}+CtrlN~\cite{zhang2023adding}-V$^\star$& \color[HTML]{9b9b9b}1 &  \color[HTML]{9b9b9b}0.479& \color[HTML]{9b9b9b}28.00& \color[HTML]{9b9b9b}0.461& &  \color[HTML]{9b9b9b}66.81& & \color[HTML]{9b9b9b}666.27 & \\
\color[HTML]{9b9b9b}Disco$^\ddagger$~\cite{wang2023disco} &\color[HTML]{9b9b9b}1&  \color[HTML]{9b9b9b}0.674& \color[HTML]{9b9b9b}29.15 & \color[HTML]{9b9b9b}0.285 & & \color[HTML]{9b9b9b}28.31 &  & \color[HTML]{9b9b9b}267.75 & \\
\color[HTML]{9b9b9b}MagicAnimate$^\star$~\cite{xu2023magicanimate} & \color[HTML]{9b9b9b}1  & \color[HTML]{9b9b9b}0.714 & \color[HTML]{9b9b9b}29.16 & \color[HTML]{9b9b9b}0.239 & \color[HTML]{9b9b9b}&\color[HTML]{9b9b9b}32.09 & & \color[HTML]{9b9b9b}179.07\\ 

\midrule
DisCo~\cite{wang2023disco}  & 1 & 0.704
& 15.16 & 0.358 & 9.11E-3
 & 76.40 & 0.827 & 689.23 & 0.908\\
MagicAnimate~\cite{xu2023magicanimate} & 1
 &  0.756 & 17.95 & 0.265 & 5.48E-3  & 57.60 & 0.846 & 374.40& 0.918\\ 
\midrule
Disco$\dagger$~\cite{wang2023disco} & 8 & 0.683 & 15.33 & 0.371 & 8.86E-3 & 65.73 & 0.813 & 807.35& 0.886 \\ 
ControlVideo~\cite{zhao2023controlvideo} & 8 & 	0.738 & 16.93 & 0.311 & 6.78E-3 & 56.50 & 0.827 & 489.30 &	0.912
  \\ 
PoseCrafter (\textbf{ours}) & 8
& 0.765
 & 17.36 & 0.275 & 5.76E-3 & 48.09
& 0.840 & 440.49& 0.921\\
\midrule
PoseCrafter (\textbf{ours}) & 16 &  0.776 & 17.87 & 0.252 & 5.23E-3 & 42.09
& \textbf{0.864} &	397.19 & 0.919 \\

PoseCrafter (\textbf{ours}) & 32 & \textbf{0.786}
 & \textbf{18.56} & \textbf{0.233}
 & \textbf{4.05E-3} & \textbf{39.65}
& 0.854 & \textbf{362.09} & \textbf{0.922} \\
\bottomrule
\end{tabular}
}
\end{table*}

\begin{table*}[tb]
\centering
\caption{Quantitative results on TED test dataset. 
NFs represent the number of available frames for the corresponding method. Image-GT signifies that these image-level metrics are calculated between each generated image and its corresponding ground truth image. $\dagger$ represents we fine-tune corresponding methods using their tailored strategies for specific subjects. $\star$ indicates the results are directly cited from the \cite{xu2023magicanimate}.
}
\label{tab:app_quantitative_results}
\resizebox{\textwidth}{!}{
\begin{tabular}{lccccccccc} 
\toprule
\multirow{2}[3]{*}{Method} & \multirow{2}[3]{*}{NFs} & \multicolumn{4}{c}{\textbf{Image-GT}} & \multicolumn{2}{c}{\textbf{Image}} & \multicolumn{2}{c}{\textbf{Video}} \\ 
\cmidrule(lr){3-6} \cmidrule(lr){7-8} \cmidrule(lr){9-10}
 & & \textbf{SSIM}\ $\uparrow$ & \textbf{PSNR}\ $\uparrow$ & \textbf{LPIPS}\ $\downarrow$ & \textbf{MSE-P}\ $\downarrow$ & \textbf{FID}\ $\downarrow$& \textbf{CLIP-I}\ $\uparrow$ & \textbf{FVD}\  $\downarrow$ & \textbf{CLIP-T} $\uparrow$ \\
\midrule
\color[HTML]{9b9b9b}TPS$^\star$~\cite{zhao2022thin} & \color[HTML]{9b9b9b}1& & & & &  \color[HTML]{9b9b9b}86.65 & &  \color[HTML]{9b9b9b}457.02& \\
\color[HTML]{9b9b9b}MRAA$^\star$~\cite{siarohin2021motion} & \color[HTML]{9b9b9b}1& & & & &  \color[HTML]{9b9b9b}35.75 & &  \color[HTML]{9b9b9b}182.78& \\
\color[HTML]{9b9b9b}IPA~\cite{ye2023ip}+CtrlN~\cite{zhang2023adding}-V$^\star$& \color[HTML]{9b9b9b}1 & & & & &  \color[HTML]{9b9b9b}49.21& &  \color[HTML]{9b9b9b}281.42 & \\
\color[HTML]{9b9b9b}Disco$^\star$~\cite{wang2023disco}& \color[HTML]{9b9b9b}1&  & & & & \color[HTML]{9b9b9b}27.51& & \color[HTML]{9b9b9b}195.00 & \\
\color[HTML]{9b9b9b}MagicAnimate$^\star$~\cite{xu2023magicanimate} &\color[HTML]{9b9b9b}1 & & & & & \color[HTML]{9b9b9b}22.78 & & \color[HTML]{9b9b9b}131.51 & \\

\midrule
DisCo~\cite{wang2023disco}  & 1 &  0.550 & 17.22 & 0.327 & 3.20E-3& 51.57& 0.794 & 309.62 & 0.909
\\
MagicAnimate~\cite{xu2023magicanimate} & 1
 &  0.498 & 13.32 & 0.338& 1.97E-3 & 46.96 & 0.801 & 194.03& 0.912
\\ 
\midrule
Disco$\dagger$~\cite{wang2023disco} & 8 & 0.551 & 17.98 &  0.373 & 3.93E-3 & 60.56 & 0.827 & 244.62 & 0.909\\ 
ControlVideo~\cite{zhao2023controlvideo} & 8 &  0.771  & 22.32 & 0.196 & 1.62E-3 &29.01 & 0.866 & 109.36 & 0.940\\ 
PoseCrafter (\textbf{ours}) & 8
& \textbf{0.810} & \textbf{23.92} & \textbf{0.142} & \textbf{1.56E-3} & \textbf{20.40}
& \textbf{0.906} & \textbf{80.01} & \textbf{0.954}\\

\bottomrule
\end{tabular}
}
\end{table*}

\subsection{More Qualitative Results}
\label{sec:more_results}
\begin{figure}[tb]
  \centering
  \includegraphics[width=1.0\linewidth]{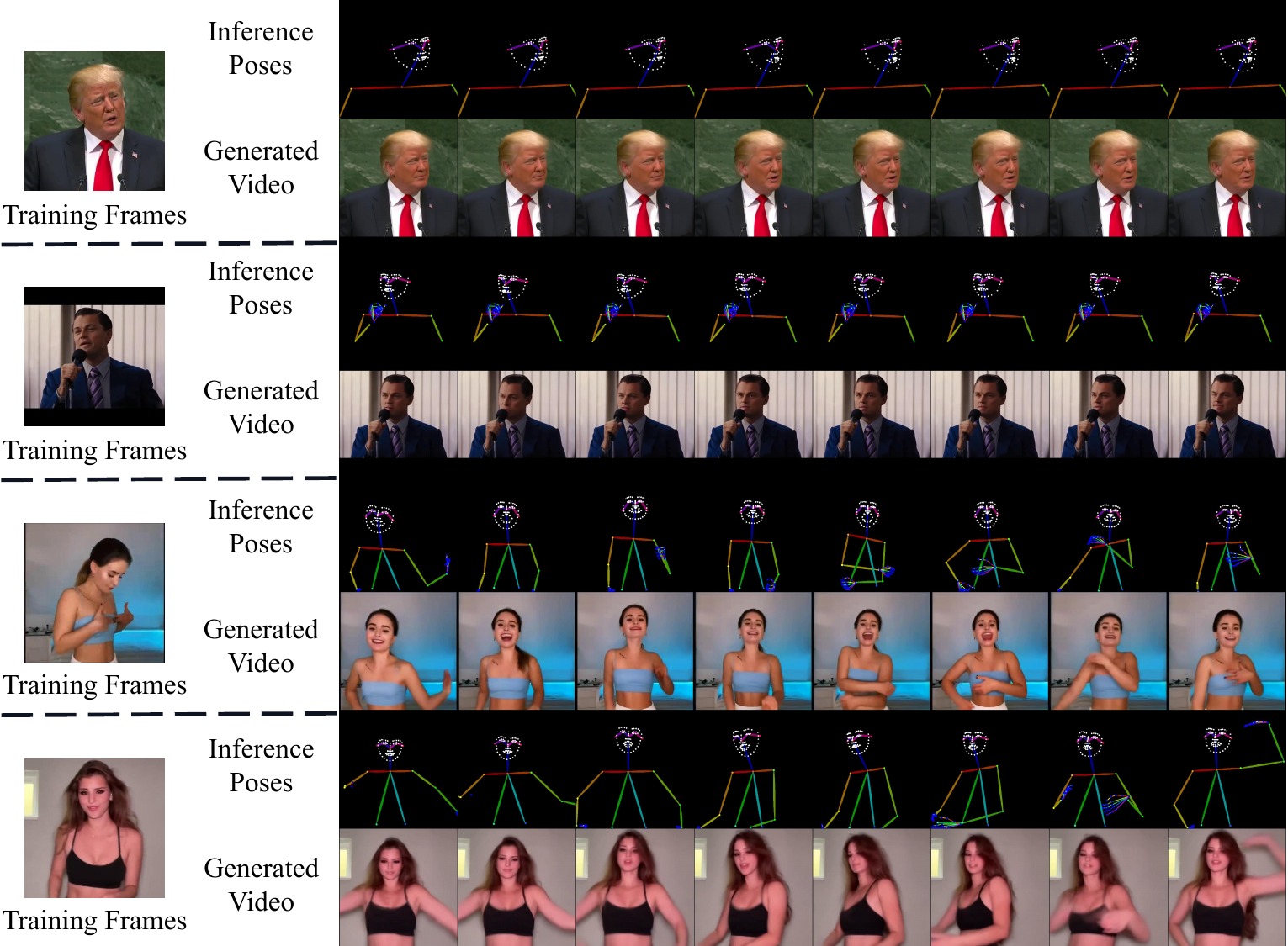}
  \caption{Inference from poses of the same individual ($N=100$ and $M=100$). Time progresses from left to right. }
  \label{fig:inference_poses_app}
\end{figure}

\begin{figure}[tb]
  \centering
  \includegraphics[width=1.0\linewidth]{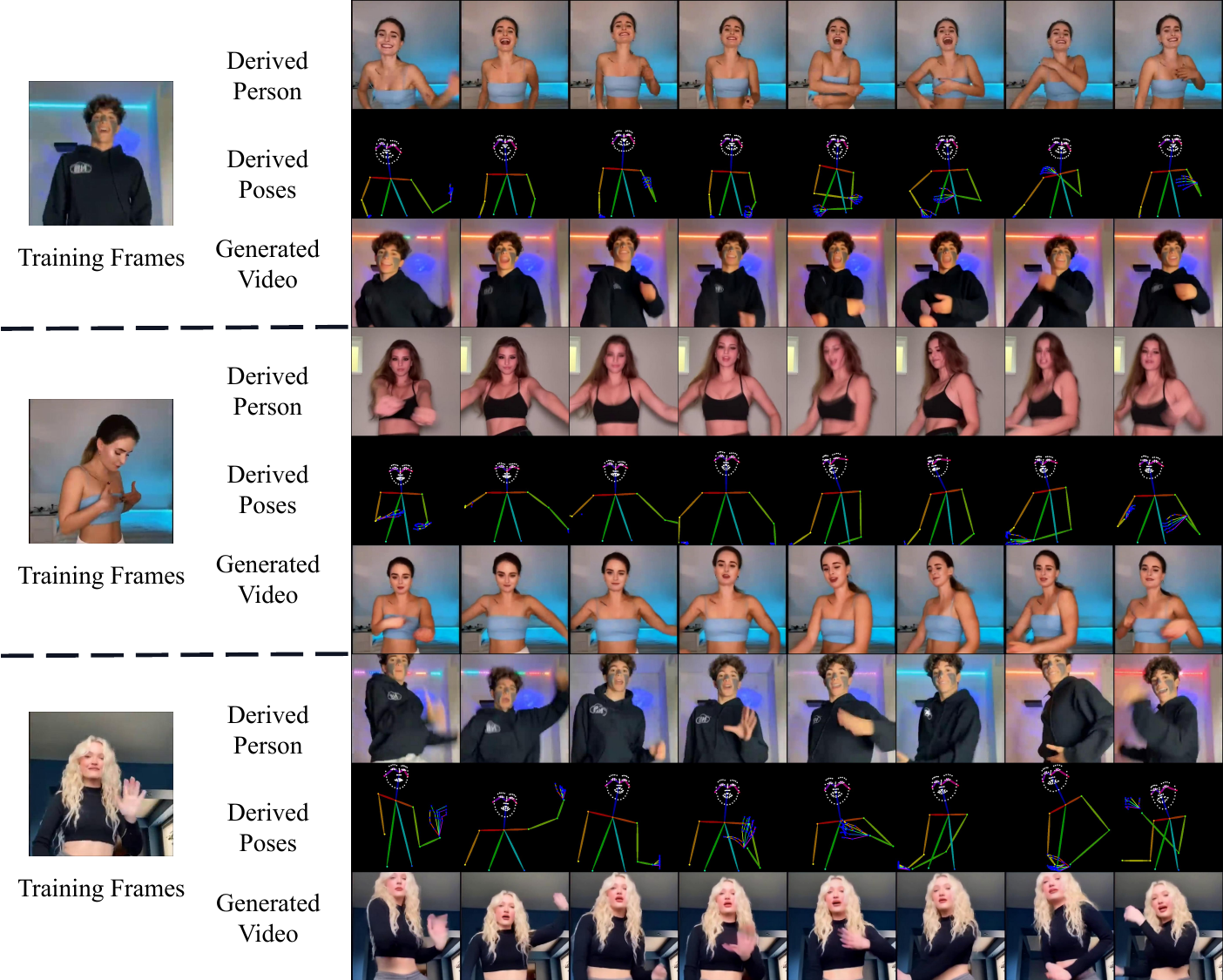}
  \caption{Inference poses of other individuals ($N=100$ and $M=100$). Time progresses from left to right. } 
  \label{fig:Inference_other_person_pose_app}
\end{figure}

\cref{fig:inference_poses_app} and \cref{fig:Inference_other_person_pose_app} show more qualitative results with the same individual poses and with poses from different individuals, respectively. Although using pose sequences with diverse motions from various humans, PoseCrafter can still maintain the human identity. In summary, PoseCrafter is capable of producing high-quality videos while allowing for flexible pose control.

\subsection{More Quantitative Results}
\label{sec:more_dataset_results}
We also conduct quantitative experiments on open-domain videos. As shown in Tab.~\ref{tab:quantitative_results_collected_dta}, consistent with conclusions on TED and TikTok, PoseCrafter excels in all quantitative metrics than all baselines on our collected open-domain dataset, demonstrating its effectiveness and robustness. 
\begin{table}
\centering
\caption{Quantitative results on our collected open-domain dataset. NFs represent the number of available frames for the corresponding method. $\dagger$ represents we fine-tune corresponding methods using their tailored strategies for specific subjects. $\ddagger$ indicates corresponding methods directly use \textbf{ground-truth} frames of target poses as input conditions.
}
\label{tab:quantitative_results_collected_dta}
\begin{tabular}{lccccc} 
\toprule
 Method & NFs & \textbf{SSIM}\ $\uparrow$ & \textbf{PSNR}\ $\uparrow$ & \textbf{LPIPS}\ $\downarrow$ & \textbf{FVD} $\downarrow$\\

\midrule
PIDM~\cite{bhunia2023person} & 1 & 0.358 & 9.36 & 0.690& 1921.43 \\
\midrule
MRAA$^\ddagger$~\cite{siarohin2021motion} & 1 & 0.654 & 15.26 & 0.423& 1101.64 \\
TPS$^\ddagger$~\cite{zhao2022thin} & 1 & 0.691 & 17.22 & 0.317 & 723.00 \\
\midrule
DisCo~\cite{wang2023disco} & 1 &  0.589 & 13.55 & 0.434 & 965.47
\\
MagicAnimate~\cite{xu2023magicanimate} & 1
 &  0.621 & 14.84 & 0.317 & 518.23
\\ 
\midrule
Disco$\dagger$~\cite{wang2023disco} & 8 &  0.638 & 15.51 &  0.310 & 583.38 \\ 
ControlVideo~\cite{zhao2023controlvideo} & 8 &  0.774  & 20.40 & 0.174 & 279.90\\ 
PoseCrafter (\textbf{ours}) & 8
& \textbf{0.808} & \textbf{21.70} & \textbf{0.143} 
& \textbf{255.61}\\

\bottomrule
\end{tabular}
\end{table}

\subsection{Failure Cases}
\label{sec:failure_cases}
\begin{figure}[tb]
  \centering
 \setlength{\abovecaptionskip}{-50pt} 
  \includegraphics[width=1.\linewidth]{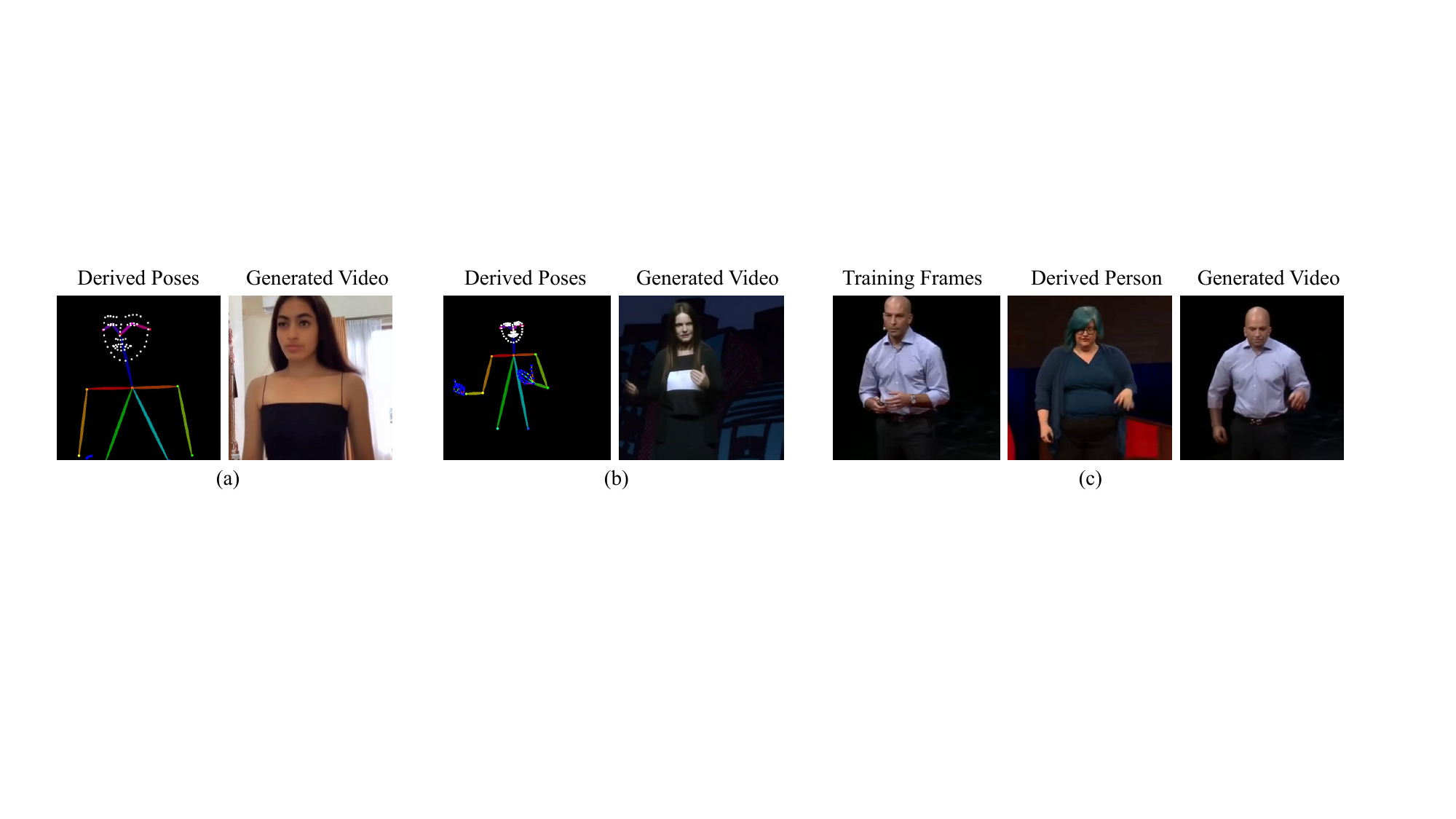}
  \caption{Failure cases include (a) misalignment of the digital human's eyes with the derived poses (i.e., closed eyes), (b) poor rendering of the right hand, and (c) changes in the facial structure of the target subject. } 
  \label{fig:failure_cases}
\end{figure}

PoseCrafter encounters several limitations in generating videos. Constrained by the capabilities of ControlNet~\cite{zhang2023adding} and the latent diffusion model~\cite{rombach2022high}, PoseCrafter produces mismatched poses (\cref{fig:failure_cases}a) and low-quality results for complex poses (\cref{fig:failure_cases}b). A significant discrepancy between training poses and inference poses causes PoseCrafter to yield low-faithfulness videos (\cref{fig:failure_cases}c), especially regarding body proportion variation. Moreover, when the training video lacks motion diversity, PoseCrafter is prone to overfitting and struggles with learning temporal consistency.

\section{Ablation}

In this section, we present a quantitative ablation study for the key designs of PoseCrafter in \cref{sec:ablation_quantitative_inference}, explore the impact of training frame quantity in \cref{sec:ablation_more_frames}, and examine the role of sample timestep choice in latent editing in \cref{sec:ablation_latent_editing}.

\subsection{Key Designs of Inference Framework}
\label{sec:ablation_quantitative_inference}

We quantitatively investigate the significance of reference frame selection, integration, and latent editing, as demonstrated in \cref{tab:ablation_for_PoseCrafter}. Specifically, reference frame selection improves all metrics compared to ControlVideo, particularly in reconstitution and quality measures, highlighting the importance of selecting an appropriate frame from the training video to DDIM inversion. Furthermore, the integration of the reference frame further enhances all metrics, confirming that putting the pose of the reference frame as the inference pose encourages similarity across it and generated frames, thereby improving video quality. In addition to enhancing the quality of hands and faces, as indicated by SSIM and PSNR, latent editing also improves temporal consistency, i.e. better CLIP-T scores.

We analyze that the features\footnote{FID and FVD use the 2D Inception and 3D Inception to extract image features, respectively.} of generated images corresponding to edited latent diverge more from real images' features than those corresponding to unedited latent, resulting in a decrease in FID and FVD scores. However, reconstruction metrics, which consider the distance between each generated frame and its corresponding ground truth, are more representative of quality than FID and FVD, which merely calculate the overall distance between all images and generated images. Hence, latent editing positively impacts video quality.

It is important to note that our strategy of selecting a reference frame from the training video whose pose coordinates are closest to inference poses may not be optimal. Therefore, exploring more effective methods for constructing a pseudo reference video is a significant area for further research.

\begin{table*}[tb]
\centering
\caption{Ablation study for key designs of the inference framework on TED. We progressively incorporate our key designs into ControlVideo and highlight both the best results and those extremely close to them in bold.}
\label{tab:ablation_for_PoseCrafter}
\resizebox{\textwidth}{!}{
\begin{tabular}{lccccccccc} 
\toprule
\multirow{2}[3]{*}{Method} & \multirow{2}[3]{*}{NFs} & \multicolumn{4}{c}{\textbf{Image-GT}} & \multicolumn{2}{c}{\textbf{Image}} & \multicolumn{2}{c}{\textbf{Video}} \\ 
\cmidrule(lr){3-6} \cmidrule(lr){7-8} \cmidrule(lr){9-10}
 & & \textbf{SSIM}\ $\uparrow$ & \textbf{PSNR}\ $\uparrow$ & \textbf{LPIPS}\ $\downarrow$ & \textbf{MSE-P}\ $\downarrow$ & \textbf{FID}\ $\downarrow$& \textbf{CLIP-I}\ $\uparrow$ & \textbf{FVD}\  $\downarrow$ & \textbf{CLIP-T} $\uparrow$ \\

\midrule
ControlVideo~\cite{zhao2023controlvideo} & 8 &  0.771  & 22.32 & 0.196 & 1.62E-3 &29.01 & 0.866 & 109.36 & 0.940\\ 
$+$ Reference-Frame Selection & 8
& 0.799 & 23.35 & 0.158 & 1.59E-3 & 22.03 & 0.892 & 74.51 & 0.942 \\
$+$ Reference-Frame Insertion & 8
& 0.802 & 23.78 & \textbf{0.141} & \textbf{1.57E-3} & \textbf{18.46} & \textbf{0.908} & \textbf{66.91} & 0.949 \\
$+$ Latent Editing & 8
& \textbf{0.810} & \textbf{23.92} & \textbf{0.142} & \textbf{1.56E-3} & 20.40 & \textbf{0.906} & 80.01 & \textbf{0.954} \\
\bottomrule
\end{tabular}
}
\end{table*}

\subsection{Training on More Frames} 
\label{sec:ablation_more_frames}

We explore the impact of varying the training video length $N$ on generated videos. For this purpose, we present Video-SIM to evaluate the resemblance between a training video $\mathbf{X}$ and a test video $\mathbf{X}'$, which is defined as:
\begin{align}
\label{eq:video_sim}
\sum_{l=1}^M \min_{1 \leq i \leq N} ||\mathbf{p}_l - \mathbf{p}_i||_2^2,
\end{align}
where $\mathbf{p}_l$ is a pose of a certain frame in $\mathbf{X}'$, $\{\mathbf{p}_i\}_{i=1}^N$ is a pose sequence of $\mathbf{X}$, and the test video length $M$ is 100 in our experiments. Essentially, Video-SIM sums up the shortest distances between each pose in $\mathbf{X}'$ and the nearest pose in $\mathbf{X}$. 

\begin{figure}[tb]
  \centering
  \begin{subfigure}{0.49\linewidth}
  \includegraphics[width=1.0\linewidth]{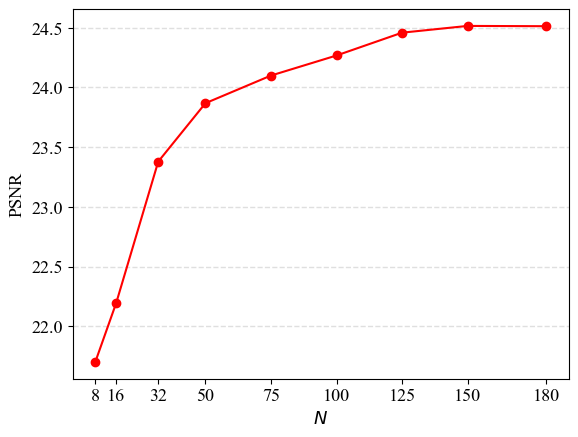}
    \caption{PSNR}
    \label{fig:short-a}
  \end{subfigure}
  \hfill
  \begin{subfigure}{0.49\linewidth}
    \label{fig:short-b}
  \includegraphics[width=1.0\linewidth]{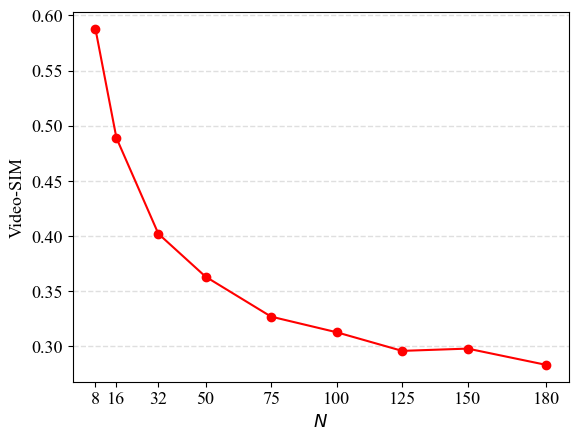}
 \caption{Video-SIM}
  \end{subfigure}

  \caption{The curves of PSNR and Video-SIM.}
  \label{fig:ablation_num_frames}
\end{figure}

\cref{fig:ablation_num_frames} depicts curves of PSNR and Video-SIM where the number of training frames ranges from 8 to 180 uniformly selected from our collected open-domain videos. With the increase in training frames, better Video-SIM enhances the similarity of generated videos to the ground truth, as reflected in higher PSNR. However, in the later phases, despite significant gains in the number of training frames, PSNR has only slight fluctuations limited by marginal improvements in Video-SIM. This suggests that gathering videos encompassing a wider range of motions pertinent to target scenes can enhance the performance of PoseCrafter.

\subsection{Implementation Time of Latent Editing}
\label{sec:ablation_latent_editing}

\begin{figure}[tb]
  \centering
  \includegraphics[width=.5\linewidth]{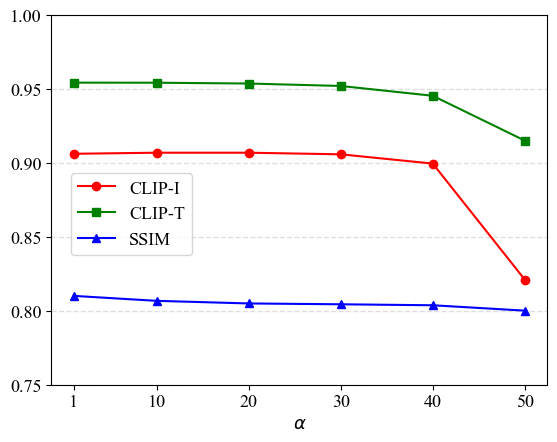}
  \caption{The curves of CLIP-T, CLIP-I, and SSIM.}
  \label{fig:ablation_alpha}
\end{figure}

We introduce a parameter $\alpha$ to determine the specific sampling time at which the latent editing operation will be executed, and we analyze the effect of $\alpha$ on generated videos in terms of CLIP-I (faithfulness), CLIP-T (temporal consistency), and SSIM (quality). As depicted in \cref{fig:ablation_alpha}, with an increase in $\alpha$, the faithfulness of generated videos first slightly rises and then decreases after $\alpha=20$. Simultaneously, temporal consistency and overall video quality consistently decrease for $\alpha$ values ranging from 1 to 50. Therefore, we set $\alpha=1$ as our default value, indicating that we edit the initial latent $\mathbf{Z}_T$ before the DDIM sampling process.

\end{document}